# Gradient-based Camera Exposure Control for Outdoor Mobile Platforms

Inwook Shim, Tae-Hyun Oh, Joon-Young Lee, Jinwook Choi, Dong-Geol Choi†, and In So Kweon

*Abstract*—We introduce a novel method to automatically adjust camera exposure for image processing and computer vision applications on mobile robot platforms. Because most image processing algorithms rely heavily on low-level image features that are based mainly on local gradient information, we consider that gradient quantity can determine the proper exposure level, allowing a camera to capture the important image features in a manner robust to illumination conditions. We then extend this concept to a multi-camera system and present a new control algorithm to achieve both brightness consistency between adjacent cameras and a proper exposure level for each camera. We implement our prototype system with off-the-shelf machine-vision cameras and demonstrate the effectiveness of the proposed algorithms on practical applications, including pedestrian detection, visual odometry, surround-view imaging, panoramic imaging and stereo matching.

## I. INTRODUCTION

Recent improvements in image processing and computer vision technologies for object detection, recognition and tracking have enabled various vision systems to operate autonomously, leading to the feasibility of autonomous mobile platforms [1]. In such real-time vision-based systems, images captured from a camera are fed directly into subsequent algorithms as input. The quality of the captured images strongly affects the algorithms' success; however, research on camera control for robust image capture has been neglected; camera control has been far less studied than have the computer vision algorithms themselves.

Most mobile platform vision systems rely on a standard auto-exposure method[1] built into the camera or by a fixed exposure hand-tuned by users. Conventional auto-exposure methods adjust camera exposure by evaluating the average brightness of an image [2, 3, 4]. Using this simple approach, undesirably exposed images are common—particularly when a scene has a significant illumination gap between dynamic ranges of the region of interest and the background. This common condition degrades the performance of the subsequent computer vision algorithms. Therefore, overcoming the problem of diverse and challenging illumination conditions at the image capture stage is an essential prerequisite for developing robust vision systems.

*This work was supported by the National Research Foundation of Korea grant funded by the Korea government under Grant 2010-0028680 and Hyundai Motor Group. (†Corresponding author:Dong-Geol Choi.)*
I. Shim is with the Ground Autonomy Laboratory, Agency for Defense Development, Korea (e-mail: inugi00@gmail.ac.kr).
T-H. Oh is with the CSAIL, MIT, USA (e-mail: thoh.mit.edu@gmail.com).
J-Y. Lee is with the Adobe Research, USA (e-mail: jolee@adobe.com)
D.-G. Choi and I.S. Kweon is with Department of Electrical Engineering, KAIST, Korea (e-mail: {dgchoi, iskweon}@kaist.ac.kr).
J. Choi is with Hyundai Motor Group, Korea.

[1]Specific methodologies are often not disclosed or are considered confidential by camera manufacturers.

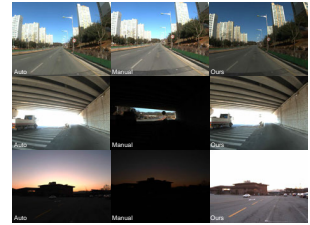

Fig. 1: Images captured under different illumination conditions. From left to right the images are from cameras with a built-in auto-exposure method, a manually tuned fixed exposure setting, and our method. Both the built-in auto-exposure method and the manual setting fail to capture well-exposed images, while our method captures images that are suitable for processing with computer vision algorithms.

More specifically, Figure 1 shows some comparisons of images resulting from the standard built-in auto-exposure and the fixed-exposure approaches in an outdoor environment. When the dynamic range of the scene is relatively narrow, both methods capture well-exposed images. Consequently, under a narrow dynamic range, a single representative parameter can easily be determined. In contrast, both methods result in undesirably exposed images under abruptly varying illumination conditions. The rationales behind these results can be characterized as follows: 1) the auto-exposure control algorithms have limited adaptability (*i.e.*, prediction), 2) do not consider the limited dynamic range of the camera, and 3) use weak criteria to assess the exposure status. We address the first and the second issues using a simulation-based approach and the third issue using a gradient-based metric.

In this paper, we present a new method to automatically adjust camera exposure using the gradient information. To handle severe illumination changes and a wide dynamic range of scene radiance, we simulate the proper exposure of the scene in the gradient domain; this process is followed by a feedback mechanism to perform auto-exposure. Because the gradient domain is robust against illumination changes and has been leveraged by many computer vision algorithms, the proposed method is suitable for capturing well-exposed images with enriched image features that are beneficial for computer

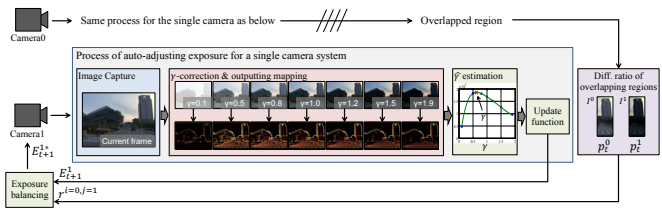

Fig. 2: The overall framework of our camera exposure control. Our method adjusts camera exposures to maximize the gradient information of the captured images. We apply a $\gamma$ correction technique to simulate information changes by exposure variations and then update the exposure settings using a real-time feedback system. In addition, we balance the exposures of multiple cameras by analyzing the intensity differences between neighboring cameras.

vision algorithms.

Figure 2 shows the overall framework of our proposed method. To build a real-time feedback control system, we use a $\gamma$ correction technique [5] that imitates exposure changes and then determines the best $\gamma$ value. Our framework implicitly guides the direction of exposure updates in the feedback system to maximize the gradient information. Moreover, in the multi-camera case, our method adjusts camera exposure by also considering neighboring cameras.

In Section II we review the existing alternatives, including both direct exposure control and post-processing; however, we stress that none of these alternatives considers the performance of subsequent algorithms, which may benefit from enriched gradients in images. In Section III-A, we describe the development of a gradient-based criterion, based on which a simulation-based control mechanism is described in Section III-B. We extend our prevision work [6] by 1) improving the stability of exposure updates in Section III-C, 2) extending the proposed methodology to a multi-camera system in Section IV, and 3) providing further technical implementation details and comprehensively analyzing the behavior of the proposed method in Section V. Finally, we conclude this work with a discussion in Section VI. Our contributions are summarized as follows:

- We propose a novel camera exposure control mechanism that leverages image gradient information to determine a proper exposure level. Specifically, we apply the novel notion of simulation-based exposure prediction, whereby the camera exposure state is updated through a newly proposed control function. This algorithm features rapid convergence to the target exposure level.
- We extend the approach for multi-camera setup, where exposure balancing issue across cameras arises.
- We implement a mobile system with synchronized cameras to prove the concept of the proposed methodology and to conduct fully controllable and ease experiments. We validate our method by extensive experiments with practical computer vision applications.

## II. RELATED WORK

Several methods exist for controlling the amount of light reaching a camera sensor: adjusting shutter speed, aperture size or gain (ISO) [7], mounting some type of density filter [8, 9, 10], or designing new camera concepts such as computational cameras [11]. Each approach has advantages and disadvantages. Shutter speed affects image blur, gain is related to image noise, and aperture size changes the depth of field. The other methods listed above require external devices, such as a density filter or a new optical device. Among these possible approaches, adjusting the shutter speed and the gain are the most popular and desirable in vision applications for mobile platforms since they ideally preserve the linear relationship between irradiance and intensity measurement. Although methods such as changing aperture size or mounting a density filter could be good complements in some beneficial situations, they can introduce artifacts[2] by changing the depth of field (causing spatially varying blur) or causing color shifts, respectively. These approaches introduce non-linear artifacts into the subsequent vision algorithms. In this paper, we focus our review on approaches that control shutter speed and gain and compare their criteria to determine the proper parameters. We then discuss algorithmic efforts to overcome the limitations of exposure control and the extensions to jointly adjust multiple cameras.

One conventional approach to achieve auto-exposure is to extract image statistics and control the camera parameters to satisfy the statistics of certain conditions. The simplest approach is to measure the average brightness over an entire image or in specific regions and adjust it to a midrange (*e.g.*, 128 for 8-bit images) [4, 12].[3] This intensity-based method may result in a proper exposure level for scenes in which illumination intensities are well-distributed across all regions and where the dynamic range of the scene is narrow. However, in practice, a scene has multiple intensity distributions over an image; thus, this simple method may lead to under- or over-saturation of important image details. To overcome this limitation, various attempts have adopted other measures that may be more robust to scene illumination variations, such as entropy [13] and histograms [14, 15]. In this paper, we maximize the sum of the log response of image gradients to enrich the image gradient information.

Several methods measure statistics over regions of interest (ROI) to create algorithms that are robust to scene variations [14, 16, 17, 18]. These methods measure the statistics of an ROI such as the image's center [12], a moving object or facial regions [18], of a user predefined ROI [14], or of foreground regions, by separating the background [16] or backlit regions [17]. However, because these methods control the camera exposure based on specific interest areas in an image, they are preferable only in specific individual applications rather than in general scenarios. Nonetheless, ROI-based methods are potentially useful because they can be combined and extended, allowing multiple measures to be adopted in a similar framework. Here, we adopt the ROI approach to assess suitable exposure levels over only interesting regions for the surround-view imaging application, as described in Section V.

---
[2]Specialized hardware implementation is required as in [8, 9, 10].
[3]Concepts of camera metering and auto exposure are introduced in: http://www.cambridgeincolour.com/tutorials/camera-metering.htm.

Other methods exploit prior knowledge of a scene to determine the proper camera exposure. Given a pre-defined reference area (black and white, brightness, or color patterns) in captured images, the exposure is adjusted based on an intensity histogram [19], color histogram [15], and image entropy [13]. While approaches based on prior information work well in a known environment, it is difficult to assume that prior information is always present in any scene—especially in most image processing applications for outdoor mobile platforms. Therefore, we focus on measuring goodness-of-fit for exposure with no reference.

The work of Zhang *et al.* is closely related to our method [20] because it mainly focuses on improving the performance of the visual odometry application and adopts a weighted version of gradient information. The primary difference lies their method for updating exposure, which is derived from the derivative of the gradient magnitude. To derive a direct update rule from the gradient measurement to exposure, they rely highly on radiometric calibration. Our method does not require radiometric calibration. Instead, we directly update the exposure based on a simulation process.

To overcome the limitations of hardware and exposure control, high dynamic range (HDR) imaging provides a way to obtain well-exposed images that are both visually and physically compelling (*i.e.*, recovering irradiance). Because all the difficulties of auto-exposure stem from the limited dynamic range of camera sensors, combining multi-exposure images can be an alternative, typical HDR approach [21, 22]. Maximizing the visual information in an HDR result requires a set of images with different exposures [23, 24]. However, such multi-exposure bracketing approaches are not suitable for dynamic scenes. While HDR methods for dynamic scenes exist [25, 26, 27], recovering the true irradiance of the dynamic portion is challenging when the inputs are captured at different times [28]. Even when a pair of low dynamic images from a stereo camera are captured simultaneously to generate an HDR image [29], the images may still suffer from misalignment artifacts due to imperfect stereo matching. This problem led to the development of a practical hardware implementation for HDR capture [8] that simultaneously captures several different exposure images of the same view. In this work, rather than adopting an expensive specialized hardware setup, we develop a method to obtain well-exposed images from off-the-shelf cameras.

Previous studies have mostly relied on single camera setup, while multi-camera exposure control has rarely been investigated. In practical image processing applications for mobile platforms, multi-camera setups are popular, and acquiring images with similar brightness between cameras is required in many applications. Nonetheless, exposure differences between cameras is common and such differences may be quite large due to different viewpoints when exposure control is applied to each camera independently. Conventional multi-camera vision applications have been developed with specially designed algorithms to compensate or overcome brightness differences. These algorithms include correspondence matching [30], panoramic imaging [31] and visual odometry [32, 33], *etc*. However, all these approaches are applied as postprocessing steps that occur after image acquisition. Unavoidably, scene information beyond the dynamic range of a camera is difficult to recover after an image has been captured; therefore, the working range of postprocessing approaches is limited by the input. In contrast, our extension for multi-camera setups allows all the cameras to obtain coherently exposed images; thus, subsequent image processing algorithms benefit from this image stability and robustness.

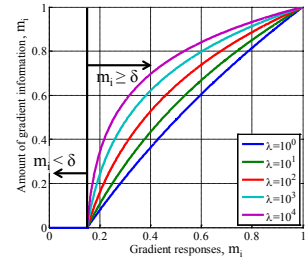

Fig. 3: Our mapping function, Eq. (1) between the gradient magnitude and amount of gradient information according to the control parameters, $\delta$ and $\lambda$. We design the mapping function by analyzing general gradient characteristics.

## III. GRADIENT BASED AUTOMATIC EXPOSURE CONTROL

### A. Image quality from an image-processing perspective

Intensity gradient is one of the most important cues for processing and understanding images. Most image features such as edges, corners, SIFT [34], and HOG [35], leverage the robustness of the intensity gradient to illumination changes. Moreover, the gradient information also characterizes object appearance and shape well; consequently, it is typically exploited in applications requiring a mid-level understanding of image content (*e.g.*, object detection, tracking, recognition, and simultaneous localization and mapping (SLAM)). Therefore, capturing images with rich gradient information is an important first step in the success of many vision algorithms. In this section, we explain how the quality of exposure of a scene is evaluated in terms of gradient quantity.

To evaluate image quality from a computer vision perspective, it is natural to exploit the gradient domain since gradient is the dominant source of visual information for many computer vision algorithms [36]. We define well-exposed images as those that have rich gradient information; therefore, we evaluate exposure by the total amount of gradient information in an image.

The gradient magnitude of a natural scene has a heavy-tailed distribution [37, 38]; consequently, most gradients have small values (including noise or zeros) relative to the maximum gradient value, and only sparse strong gradients exist. Because these strong gradients typically occur around object boundaries, they have a high probability of embedding important information. On the other hand, gradients are sensitive to subtle intensity variations (*i.e.*, small magnitude gradients such as image noise should be filtered properly).

To balance the importance of weak and strong gradients, we use a logarithmic function[4] to map the relationship between gradient magnitude and the quantity of gradient information. However, logarithmic mapping may over-emphasize small gradients that occur due to image noise. To perform mapping that is robust against image noise, we modify the mapping function to discard small noise values using a simple threshold, defined as follows:

$$\bar{m}_i = \begin{cases} \frac{1}{N} log(\lambda(m_i - \delta) + 1) & \text{for} \quad m_i \geq \delta \\ 0 & \text{for} \quad m_i < \delta \end{cases} \quad (1)$$

where $N = log(\lambda(1-\delta)+1)$, $m_i$[5] is the gradient magnitude at pixel location $i$, $\delta$ is the activation threshold value, $\lambda$ is a control parameter to adjust mapping tendencies, and $\bar{m}_i$ represents the amount of gradient information corresponding to gradient magnitude. In addition, $N$ is a normalization factor that restricts the output range of the function to $[0, 1]$.

---
[4]In information theory, the quantity of information is often measured by a logarithmic function, *e.g*., entropy

Eq. (1) has two user control parameters, $\delta$ and $\lambda$, that allow users to tune our method based on their needs. The parameter $\delta$ determines the activation threshold value: the mapping function regards a gradient value smaller than $\delta$ as noise and ignores it. The parameter $\lambda$ determines the mapping tendencies. We can emphasize strong intensity variations by setting $\lambda$ to a small value or emphasize subtle texture variations by setting $\lambda$ to a large value. Figure 3 shows a plot of the mapping functions with varied control parameters. Using Eq. (1), we compute the total amount of gradient information in an image as $M = \sum \bar{m}_i$. Our method regards images with larger values of $M$ as better-exposed images that contain more rich gradient information in a scene. In all the experiments in this work, we empirically set $\delta$ and $\lambda$ to 0.06 and $10^3$, respectively. The related experiments are presented in Section V-A2.

### B. Auto-adjusting camera exposure

Our method adjusts camera exposure at each frame by increasing the proposed criterion in Eq. (1). One challenge is that the true relationship between exposure time and sensed gradient (or intensity) is unknown [40]; instead, complex imaging pipelines exist in modern cameras. Revealing the effects of such pipelines is another challenging research area. Thus, rather than modeling the complex relationship explicitly, we propose an alternative approach that avoids the difficulty involved in modeling the imaging pipeline. We develop a simulation-based feedback system that allows us to find the approximate direction of the exposure to be updated.

We simply adopt a $\gamma$-mapping to simulate exposure changes, which roughly approximates non-linear imaging pipelines as well as under- or over-saturation effects. We generate $\gamma$-mapped images $I_{out} = I_{in}^\gamma$ from the current input image $I_{in}$[6]. The $\gamma$-mapping results in a darker image when $\gamma < 1$, and in a brighter image when $\gamma > 1$. Using this approach, we simulate exposure changes from a batch of $\gamma$-mapped images, compute the amount of gradient information for each image, and then find a $\gamma$ such that maximizes the gradient information:

$$\arg\max_\gamma \ M(I_{in}^\gamma). \quad (2)$$

---
[5]In this paper, the gradient magnitude is computed by the Sobel operator [39], and we normalize the magnitude to a range of [0, 1].
[6]We assume that intensity is in the range of [0, 1].

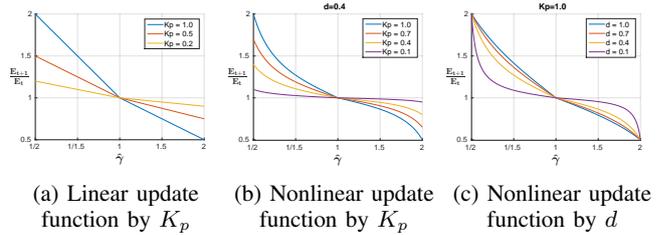

(a) Linear update function by $K_p$  (b) Nonlinear update function by $K_p$  (c) Nonlinear update function by $d$

Fig. 4: Examples of linear and nonlinear update functions with varying values of the control parameters $K_p$ and $d$.

Computing Eq. (2) for all possible $\gamma$ values would not be a negligible operation for high-resolution image regimes. However, computing the gradient information at such a fine scale may be unnecessary because exposures are insensitive to subtle gradient differences in the image domain to some extent; hence, as a trade-off between computational cost and accuracy, an effective number of anchor images and image resolution must be determined. To determine a proper option, we reveal the trade-off relationship through synthetic experiments in Section V-A3.

At runtime, to improve the accuracy of the $\hat{\gamma}$ estimate, we first compute the gradient values for each anchor image and then fit the gradient values with a fifth-order polynomial. We pick the maximum value of the polynomial function in the range of $\gamma = [\frac{1}{1.9}, 1.9]$ and assign its corresponding $\gamma$ value to $\hat{\gamma}$. The range parameters are empirically determined; the related experiments can be found in the supplementary material.

### C. Camera exposure update function

In this section, we describe the update rules for camera exposure given $\hat{\gamma}$. The update rules are designed such that the current exposure will move toward a value that eventually results in $\hat{\gamma}$ being 1. We propose two methods to update the camera exposure: a linear update function and its extension to a nonlinear version. These two functions are designed to adjust camera exposure in inverse proportion to $\hat{\gamma}$. Figure 4 (a–c) shows how the linear and nonlinear update functions work based on the control parameters, $K_p$ and $d$. The details of the update functions with the control parameters are described as follows.

**Linear update function.** The linear update is defined as

$$E_{t+1} = (1 + \alpha K_p(1-\hat{\gamma}))E_t, \ \alpha = \begin{cases} 1/2 & \text{for} \quad \hat{\gamma} \geq 1 \\ 1 & \text{for} \quad \hat{\gamma} < 1, \end{cases} \quad (3)$$

where $E_t$ is the exposure level at time $t$, and $K_p$ is the proportional gain that controls the convergence speed by adjusting the maximum and minimum values of the exposure update ratio. Figure 4 (a) shows the tendency of this function according to $K_p$. As shown in the figure, there is a trade-off between the convergence speed and the stability of the update function. A high $K_p$ value causes the feedback system to catch up quickly but may cause oscillation and overshooting.

**Nonlinear update function.** The oscillation and overshooting problems of the linear update function are caused by the non-smooth transition of exposure update ratios at the convergence point ($\hat{\gamma} = 1.0$) (the curved point in Figure 4 (a)). To relieve this problem, we designed a new update function that is parametric and nonlinear. This nonlinear update function is defined as follows:

$$E_{t+1} = (1 + \alpha K_p(R-1))E_t$$
$$\text{s.t.} \quad R = d \cdot \tan\{(2-\hat{\gamma}) \cdot \arctan(\tfrac{1}{d}) - \arctan(\tfrac{1}{d})\} + 1, \quad (4)$$

where the same $\alpha$ with Eq. (3) is used. The $R$ mapping is designed to realize curved shapes. The nonlinear update function has two control parameters $K_p$ and $d$. Figure 4 (b) and (c) show how the parameters $K_p$ and $d$ work to control the convergence speed and stability of exposure update ratios. As with the linear update function, $K_p$ controls the speed of convergence from the current exposure level to the desired exposure level. The additional control parameter $d$ controls the nonlinearity of the update function. As shown in the figure, a smaller $d$ value causes greater nonlinearity of the slope of the update ratios, which results in a smoother transition at the convergence point.

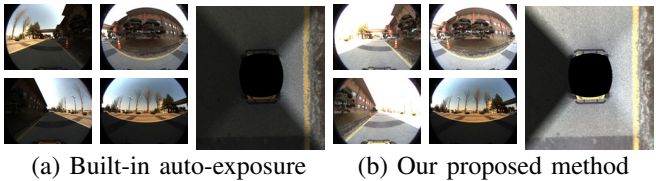

(a) Built-in auto-exposure    (b) Our proposed method

Fig. 5: The images in this figure show one of the worst cases in the performance of the surround view application caused by low inter-image consistency. Images (a) and (b) show the original images and the result of the surround-view image using the built-in camera auto-exposure and the proposed method, respectively. Each camera is individually adjusted without exposure balancing.

## IV. EXPOSURE BALANCING FOR MULTI-CAMERA SYSTEM

Figure 5 shows an example of a surround-view imaging application that stitches multiple camera images together. Even though the images captured by each camera have a reasonable quality, the brightness inconsistency among the images degrades the overall quality of the surround-view image. This type of problem in multi-camera systems can easily be observed in many applications, such as 360° panoramic imaging and multi-camera tracking when each camera controls its exposure individually. In contrast, when all the cameras share the same camera exposure to achieve constant brightness levels, the system captures only a narrow dynamic range and may lose important scene information.

Using our gradient-based exposure control approach, we also address the problem of advanced exposure control for multi-camera systems, which are popular in many vision systems such as autonomous vehicles. While independent control of each camera enables capturing of better image features, many cases exist for multi-view image processing such as panoramic imaging and stereo matching that favor maintaining consistent brightness among adjacent cameras. To address such cases, we present an exposure balancing algorithm that both

**Algorithm 1:** Gradient-based Camera Exposure Control

1 function Main ($camera_i, camera_j = Null$);
  **Input** : Camera index (i) and its neighboring cameras (j)
2 Set Camera Mode($camera_i$, built-in $AE$);
3 pause(1);
4 Set Camera Mode($camera_i$, manual);
5 **while** $camera_i$ is alive **do**
6 | $E_i \leftarrow$ Get Exposure ($camera_i$);
7 | $I_i \leftarrow$ Get Image ($camera_i$);
8 | **for** $\gamma$ in anchors **do**
9 | | $I_i^\gamma \leftarrow$ Gamma Correction ($I_i, \gamma$);
10 | | $M_i^\gamma \leftarrow$ Compute Gradient ($I_i^\gamma$), Eq. (1);
11 | **end**
12 | $\hat{\gamma} \leftarrow argmax_\gamma M_i^\gamma$, Eq. (2);
13 | $newE_i \leftarrow$ Update Exposure ($\hat{\gamma}, E_i$), Eq. (3 or 4);
14 | **if** $camera_j$ is not Null **then**
15 | | $I_j \leftarrow$ Get Image ($camera_j$);
16 | | $newE_i \leftarrow$ Balancing ($I_i, I_j, E_i, newE_i$), Eq. (5);
17 | **end**
18 | Set Exposure ($camera_i, newE_i$);
19 **end**

achieves gradient-based camera control and simultaneously maintains brightness consistency between neighboring images.

On top of our gradient-based exposure control, we formulate an optimization problem that determines camera exposure by considering a balance between individual image quality and brightness constancy. Given the desired camera exposures obtained by Eq. (4), our optimization function considers the desired exposure as a unary term and brightness similarity in overlapped regions as a pairwise term. The optimization function is defined as follows:

$$E_{t+1}^{i*} = \arg\min_X \quad \alpha^i \cdot \mathcal{E}_u(i,t) + \tfrac{1-\alpha^i}{N} \sum_{j \in G(i)} \mathcal{E}_p(i,j,t),$$
$$\text{s.t.} \quad \mathcal{E}_u(i,t) = \|X - E_{t+1}^i\|^2,$$
$$\mathcal{E}_p(i,j,t) = \|X - r^{ij} \cdot E_t^{i*}\|^2,$$
$$r^{ij} = median\left(\tfrac{mean(p^j)}{mean(p^i)+\epsilon}\right), \quad p \in P,$$
(5)

where $i$ denotes a target camera on which the exposure level should be updated, $j$ denotes a neighboring camera whose field of view overlaps with camera $i$, $N$ is the number of neighboring cameras, and $t$ denotes the time sequence. Here, $E_{t+1}^{i*}$ indicates an estimated optimal exposure of camera $i$ for the $t+1$ frame (next frame), and $E_{t+1}^i$ is the estimated exposure value of camera $i$ from Eq. (4). $P$ represents a set of overlapped patches between two cameras, $mean(p)$ is the average patch intensity, and $r^{ij}$ is the median value of the relative brightness ratios between corresponding patches from cameras $i$ and $j$. We use the $median$ ratio of the average patches to make our solution robust to image noise and misalignment of the overlapped regions.

The exposure values among cameras were balanced by our exposure balancing method in Eq. (5). The control parameter $\alpha$ in Eq. (5) adjusts the weights of the unary and pairwise terms. To account for varying illumination conditions well, we updated $\alpha$ as follows:

$$\alpha^i = \begin{cases} (1-R^i) + 0.5 & \text{if} \quad R^i < 1.0, \\ R^i/2 & \text{if} \quad R^i \geq 1.0, \end{cases} \quad (6)$$

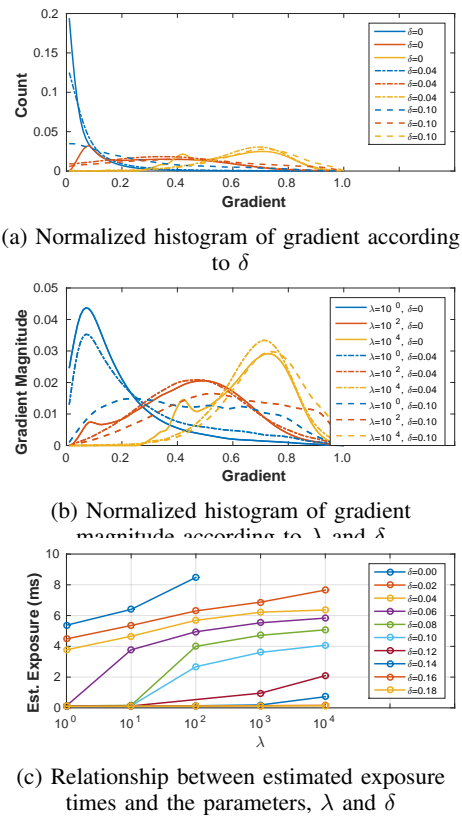

(a) Normalized histogram of gradient according to $\delta$

(b) Normalized histogram of gradient magnitude according to $\lambda$ and $\delta$

(c) Relationship between estimated exposure times and the parameters, $\lambda$ and $\delta$

Fig. 6: Changes of gradient distributions according to the $\lambda$ and $\delta$ parameters: (a) normalized histogram of the gradient by $\delta$, (b) distributions of gradient magnitude by $\lambda$ and $\delta$, and (c) the relationship between desired exposure times and the user parameters. Additional experimental results can be found in the supplementary material.

where $i$ is the camera index, and $R$ is the exposure ratio between the current and estimated exposure values of our update function as $R^i = \frac{E^i_{t+1}}{E^i_t} \in [0.5, 2.0]$. Therefore, we give a large weight to the unary term for fast scene adaptation when the exposure levels vary widely, and we increase the influence of the pairwise term to quickly converge to an exposure balancing point when the exposure levels are largely stationary.

In this optimization, $\mathcal{E}_u(\cdot)$ and $\mathcal{E}_p(\cdot)$ denote the unary and pairwise terms, respectively, and the two terms are balanced by $\alpha$. Therefore, a large $\alpha$ encourages more gradient information in a scene while a small $\alpha$ enforces the brightness similarity between cameras. Eq. (4) has a closed-form solution. For each frame, we solve Eq. (4) and Eq. (5) only once for each camera $i$ because they will eventually converge through our feedback system. This procedure allows the exposure parameters to progressively adapt to a scene. Algorithm 1 describes the entire process of the proposed method.

## V. EXPERIMENTAL RESULTS

To validate the performances of our proposed methods, we conduct both synthetic experiments and real experiments. In the synthetic experiments, we analyze the various components of our proposed method regarding image quality metric, exposure estimation, and the update functions described in Section III-A, Section III-B, and Section III-C, respectively. In the real experiments, we show the effectiveness of our proposed method with five image processing applications. The surveillance and automotive visual odometry experiments demonstrate the results of the proposed single-camera exposure control method (Ours-single, henceforth). We also conduct experiments using surround-view, panoramic imaging, and stereo matching applications using the proposed multi-camera exposure control method (Ours-multi, henceforth). We compare our method to two conventional methods, a camera built-in auto-exposure method (AE, henceforth) and a manually tuned fixed exposure setting (ME, henceforth). We perform all the experiments using machine vision cameras that have a linear camera response function and configure the initial camera exposures using the AE method. All the parameters in our method are fixed for all the experiments. You can find additional materials such as video clips on our project webpage[7].

### A. Analysis of our proposed method

*1) Synthetic experiments using HDR:* We devise a synthetic simulator-based experiment to facilitate the complicated comparison process in reproducible test environments. We mimic a simplified camera imaging pipeline for rendering low dynamic range images (LDR) from high dynamic range images (HDR) based on desired exposure times and pre-compute the ground truth values for $\gamma$ and exposure. Given this controlled simulator, we tested various setups to observe the relationship between exposure times and our gradient metric according to the $\lambda$ and $\delta$ values in 2), the trade-off relationship between the number of anchor and image resolution in 3), and the convergence speed of the update functions in 4).

*2) Analysis of image quality metric:* We analyze the gradient statistics of a sample HDR scene according to the user parameters described in Section III-A. Figure 6 shows the gradient distributions, from which we can observe the effects of changing $\lambda$ and $\delta$ values on scene characteristics and desired exposure times: (a) shows the histogram changes resulting from the activation threshold $\delta$. A high $\delta$ value removes considerable gradient information when computing gradient magnitude and adversely affects the exposure estimation; (b) shows the gradient magnitude over the entire gradient range; (c) shows the relationship between the estimated exposure and the two parameters.

Given the results of this experiment, we empirically set $\delta$ and $\lambda$ to 0.06 and $10^3$, respectively. These values show good patterns of the gradient distributions and they work consistently on the synthetic simulation compared to other parameter values. We fix these two parameters throughout all the subsequent experiments.

*3) Performance analysis according to the number of anchors and image resolution:* Figure 7 shows the performance variations according to the number of anchors and image resolution. We measure the error against the ground truth $\gamma$, which is obtained by exploiting all the finely quantized candidate $\gamma$ values. Figure 7 (a) shows that at least five anchors are required to achieve a stable $\gamma$ estimation. The computational speed depends on both the image resolution and the number of anchors. (b) shows the processing time of the proposed method with different image resolutions, and (c) shows how the image resolution affect the accuracy of the estimated exposure. In terms of our proposed metric, we consider the exposure computed from an image resolution of $2,560 \times 1,280$ pixels as a reference.

---
[7] https://sites.google.com/site/iwshimcv/home/multicamera

In this section, we investigate the relationship between the effective number of anchor images and image resolution to understand the trade-offs between the computational cost and accuracy. To determine a proper value, we provide the trade-off relationship through synthetic experiments in Figure 7.

Given the results of this experiment, we choose seven anchors, $\gamma \in [\frac{1}{1.9}, \frac{1}{1.5}, \frac{1}{1.2}, 1.0, 1.2, 1.5, 1.9]$, and fix them through all the subsequent experiments in this work. These anchors have similar errors as other larger anchor sets but achieve a reasonable processing time. For the image resolution, we adopted a resolution of $320 \times 240$ pixels to achieve real-time performance ($14.15ms$ using an Intel Core i5-6260U@1.80GHz without parallel processing); this resolution preserves most of the major structures in the original image.

*4) Analysis of the convergence speed of the update methods:* To determine the proper parameter values for the linear and nonlinear update functions, we ran a parameter-sweeping experiment. To conduct this experiment, we control the environment in a dark room with controllable illumination for two reasons: 1) the parameters are related only to the convergence speed and stability; thus, various levels and continuous illumination variation is necessary, and 2) the parameters are invariant to the absolute value of illumination but only relevant to its changes; thus we can generalize the results to a real environment.

We used five LED light sources. Each LED was switched on and off repeatedly at different time intervals controlled by digital timers. All the sets of parameters were tested at intervals of 0.1, and we empirically evaluated the convergence speed and stability for each parameter set. Figure 8 shows the results with the best-performing parameters, which resulted in fast convergence with low overshoot and small oscillations. Due to the smooth slope of the nonlinear update function around the convergence point $(1, 1)$, the nonlinear update function rarely suffered from overshoot or oscillation, while the linear update function suffered from both problems.

We compared the proposed update methods with Zhang *et al.* [20] and standard PID [41] using the synthetic experiment described in Section V-A1. To implement the method of Zhang *et al.*,[8] intensity boundary values are used as additional control parameters to limit the operating range and avoid uncontrollable situations. When the mean intensity of the captured image is within the predefined boundaries, the method works plausibly. However, when the mean intensity is outside the boundary range, this method controls the exposure by relying solely on image intensity. Because this parameter leads to noticeable behavior differences in Zhang's method, we tested it with varying boundary parameters, *i.e.*, $(70, 190)$, $(50, 210)$, $(30, 230)$, and $(10, 250)$, in Figure 9. For example, Zhang–$(70, 190)$ controls the exposure based on its gradient metric only when the mean intensity of a captured image is between 70 and 190. In the range from the 1-st to 20-th frame in Figure 9-(top), Zhang–$(70, 190)$, –$(50, 210)$, and –$(30, 230)$ control the exposure using mean intensity only, because the given mean intensities are outside of the predefined boundaries. After that, they switch their control scheme into the gradient-based method. Zhang–$(10, 250)$ does not update the exposure, because the camera response function was unable to provide gradients able to update at low intensities from the initial frame. We also observed that Zhang *et al.* is sensitive to the quality of the radiometric calibration.

[8]We obtained the original code from the authors, and we used all the parameters suggested by the authors. The intensity boundary parameter $(70, 190)$ is the original parameter used by the authors.

We also implemented a reference method using the standard PID controller. Because we can consider that the optimal value of $\hat{\gamma}$ should be one when the exposure is desirable, we use the difference between the estimated $\hat{\gamma}$ and 1 as the input error to implement the PID-based method. We tuned the PID parameters based on the Ziegler–Nichols tuning method [41, 42]. While we attempted to further tune the PID update function and find the model that works best, we did not find parameters that achieved faster convergence than our proposed nonlinear update method with small overshoot and small oscillations. Our linear update function performs comparably to Zhang *et al.* and the PID-based method, and our nonlinear update function performs favorably against all the other methods in terms of the exposure convergence speed, while its qualitative results at the converged points are comparable (see the supplementary material).

*B. Single-camera experiments*

*1) Implementation:* Figure 10 (a) shows the camera system used to evaluate the single-camera exposure control. For comparative evaluation, we used three Flea3 cameras, each of which is equipped with a Sony ICX424 CCD $1/3''$ sensor with a $640 \times 480$ resolution and a 58.72 dB dynamic range. The three cameras are placed in parallel and synchronized by using an internal software trigger. Each camera's exposure parameters were determined by AE, ME, and Ours-single.

We used two camera parameters, *shutter speed* (exposure time) and *gain*, because controlling the shutter speed affects the frame rate, which is often critical for an application. We describe both parameters as the exposure level $E$ in Eq. (3). In our implementation, to increase $E$, we first control the shutter speed based on the value of $E$ until it reaches a pre-defined maximum value (we set it to $25.51ms$ for the following two applications, surveillance and automotive visual odometry). Then, after the shutter speed reaches the maximum value, we increase the gain. In the opposite case, the shutter speed is adjusted when it is less than the maximum value.

*2) Surveillance application:* To validate Ours-single in a surveillance application, we recorded image sequences every two hours from 8:00 to 18:00 each day. We collected two types of datasets.

**Dataset.** One dataset was collected from the three cameras after the cameras reached steady-state (SURVEILLANCE-A).

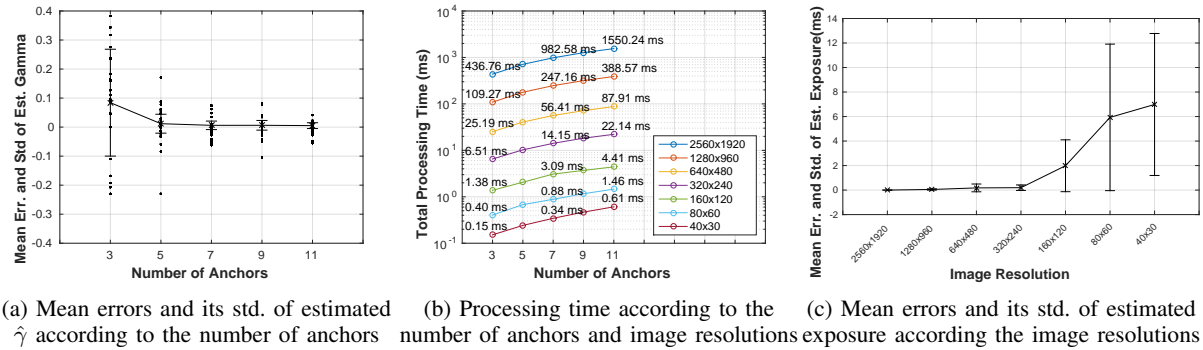

(a) Mean errors and its std. of estimated $\hat{\gamma}$ according to the number of anchors

(b) Processing time according to the number of anchors and image resolutions

(c) Mean errors and its std. of estimated exposure according the image resolutions

Fig. 7: Performance analysis based on the number of anchors and image resolutions: (a) the image resolution was fixed to $2560 \times 1920$ to test varying numbers of anchors, and (c) we used 11 anchors to analyze the effects of varying the image resolution.

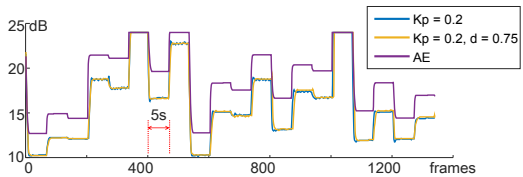

Fig. 8: Camera parameter changes according to illumination changes. The built-in auto exposure method has different exposure values than our method because it uses different exposure evaluation metrics.

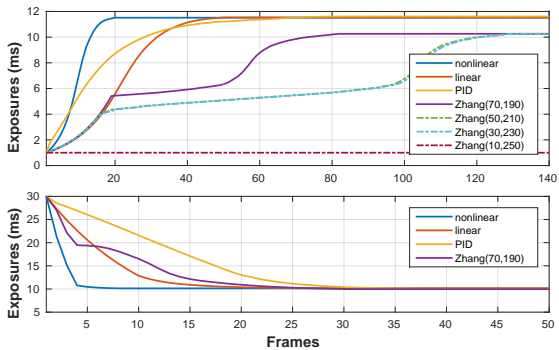

Fig. 9: Convergence speed comparison of update methods in the synthetic simulator. The top and bottom graphs show the respective scenarios under increasing and decreasing exposures. The differences between the saturation points of our methods and those of Zhang *et al.* [20] are caused by the use of different image quality metrics.

Sequences were recorded for approximately 10 minutes at every time step. This dataset was used to compare the performance of pedestrian detection as a surveillance application; approximately 2,500 pedestrians appeared in the sequences. For the ME method, we used the same parameter in the dataset, which was initially determined to be 8:00 by manual tuning.

The other dataset was acquired by sweeping the full range of possible exposure levels to validate our algorithm (SURVEILLANCE-B). We sampled 210 exposure parameters covering the full exposure range; therefore, the dataset consisted of $1,260 \ (= 210 \times 6$ time steps) images.

**Comparison of steady-state exposures.** Figure 11 shows a comparison of the results of our method with AE and ME

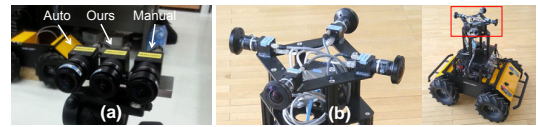

Fig. 10: Camera systems used for the experimental validations. The three cameras in (a) have the same hardware specifications, and they are synchronized using an internal software trigger. Each camera's exposure parameters were determined by AE, Ours-single, and ME, respectively. (b) shows the multi-camera system on a mobile robot (Clearpath HUSKY A200). The four-camera system has the same hardware specifications, and they are synchronized by a hardware trigger.

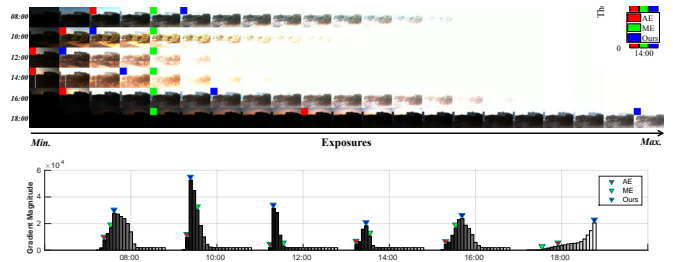

Fig. 11: Real scene comparison in the SURVEILLANCE-B dataset. The images in the top rows were captured every two hours from 8:00 to 18:00. In each row, we display 21 images sampled from 210 different exposure images at every time step. The graph at the bottom shows bar graphs of the gradient magnitudes corresponding to the 21 images at each time step. For example, the third bar in the first column (08:00) indicates the gradient magnitude of the third image in the first row of the top. The red, green, and blue markers indicate the best exposures (steady-state exposure) recommended by the respective methods at each time step.

in a real daytime scene from the SURVEILLANCE-B dataset. At each acquisition time, we annotate the color markers to indicate the best-exposed image as recommended by the respective methods. While we tuned ME against the scene at 8:00 AM, its fixed parameter obviously causes the results to suffer from illumination changes over time. The AE approach excludes textural details due to the dominant illumination from the sky region. In contrast, Ours-single properly adjusts the camera parameters to capture images with the maximum gradient information at every time step. Our method is more likely to result in an exposed image that both quantitatively and qualitatively contains richer textural details than the results of the other methods.

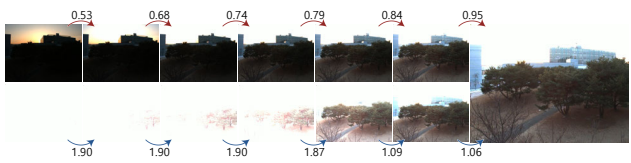

Fig. 12: Progress of our feedback system. We put the leftmost images into our feedback system, and the system iteratively converges to the rightmost image. The numbers indicate the $\hat{\gamma}$ values estimated by Ours-single.

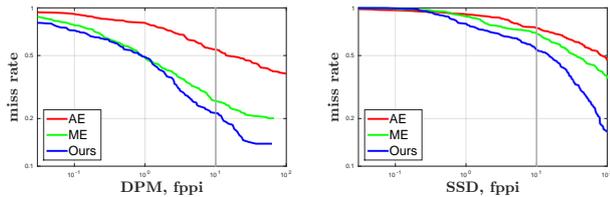

Fig. 13: Quantitative evaluation results of the pedestrian detection experiment. Results closer to the bottom left indicate better performance. The graphs show the false positive per image (fppi) of DPM [43] and SSD [44].

**Validation of our feedback system.** In Figure 12, we show the feedback system procedure, which is explained in Section III-B. To simulate extreme cases in which the illumination varies rapidly, we applied Ours-single to SURVEILLANCE-B. In the figure, the output images of our feedback system are presented from left to right. We first put both an under-exposed image (the leftmost image on the top row) and an over-exposed image (the leftmost image on the bottom row) into our feedback system. From each image, our algorithm estimated $\hat{\gamma}$ and updated the camera exposure according to Eq. (3). Using an updated camera exposure, we took an image that matched the exposure parameter in the dataset, and iteratively applied our feedback system until it converged. Our system converged to the rightmost image, which demonstrates that our method can reliably adjust camera exposure even in extreme cases. The numbers in the figure indicate the $\hat{\gamma}$ values estimated by Ours-single. We can observe that $\hat{\gamma}$ converged to one as the output images converged to the appropriate exposure level in the rightmost image.

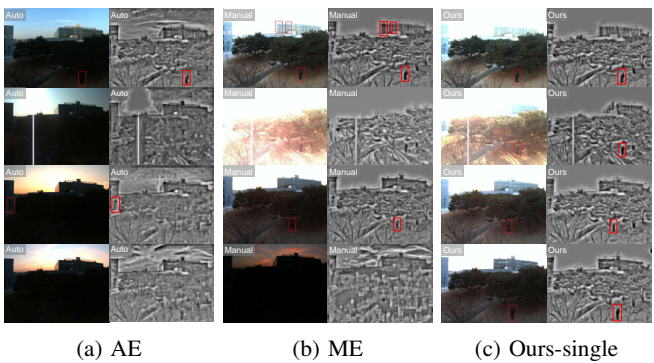

(a) AE (b) ME (c) Ours-single

Fig. 14: The images show example results of pedestrian detection and visualizations of feature spaces using HOGgles [36]. All the input images are from the SURVEILLANCE-A dataset, and the images in each row were captured at the same time. From top to bottom, the images were captured at approximately 8:00, 14:00, 16:00, and 18:00.

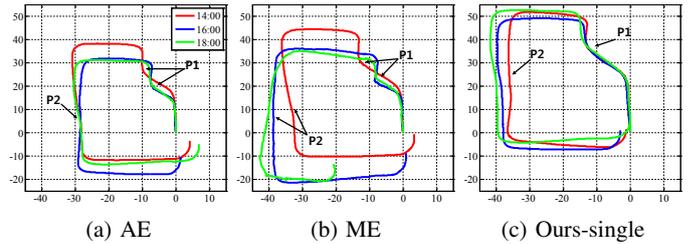

(a) AE (b) ME (c) Ours-single

Fig. 15: Trajectories estimated by the visual odometry [45]. Because the vehicle followed a homing path, the ground-truth values of the start and end points of all trajectories are centered on $(0, 0)$

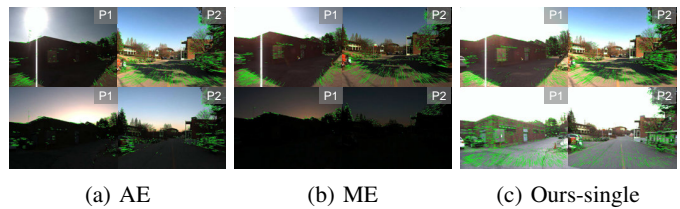

(a) AE (b) ME (c) Ours-single

Fig. 16: Images are shown from two locations on the path as indicated in Figure 15. The top row shows images at 14:00 and the bottom row shows images at 18:00. The green lines indicate tracked image features between adjacent frames.

Note that the scene has a dynamic range that exceeds the dynamic range of our camera. In the same situation, the AE method wiped out important details in low radiance areas to prevent saturation in the sky region, as shown in Figure 14. Our method naturally adjusts camera exposure to emphasize important details by evaluating camera exposure in the gradient domain.

**Pedestrian detection.** In this experiment, we compared the pedestrian detection performance with that of other exposure methods. Pedestrian detection is an important task in surveillance. The pedestrian detection sequences for this experiment came from the SURVEILLANCE-A dataset, and we used two types of detectors: hand-craft gradient feature based pedestrian detector, the DPM [43] and the convolutional neural network based object detector, SSD [44].[9]

Figure 13 shows the quantitative evaluation result of the pedestrian detection experiment. For this evaluation, the ground-truth was manually labeled for all the data. Following [48], we adopt the miss rate against false positives per image (FPPI) as an evaluation metric. The evaluation metric indicates better performance as results get closer to the bottom left; therefore, it shows that our method has better capability of preserving the details of images at the capturing stage than do the conventional AE and ME methods.

---

[9]The SSD model were pre-trained on VOC 2007 and 2012 datasets [46, 47]

Figure 14 shows some example results of pedestrian detection. The images obtained by the AE method are under-exposed due to the high illumination in the sky region, and the images obtained by the ME method are over- or under-exposed due to illumination changes. More importantly, the pedestrian detector was sensitive to image quality; it failed to detect humans in poorly exposed images.

We used HOGgles [36] to visualize images from an image processing perspective to qualitatively evaluate the exposure control algorithm. HOGgles inverts HOG feature spaces back to a natural image; therefore, it is useful for understanding how a computer sees visual images. In Figure 14, the HOGgles visualizations of our method consistently show detailed features despite large illumination changes, while the AE and ME methods were unable to preserve the visual information in low radiance regions due to incorrect exposures. The HOGgles visualization clearly demonstrates that our method is more suitable than are the conventional AE and ME methods for outdoor computer vision tasks.

*3) Automotive visual odometry application:* We performed the visual odometry experiments using the automotive driving dataset. Visual odometry is the process of incrementally estimating the pose of a vehicle by analyzing images captured by a vehicle-mounted camera. This experiment shows how the image-capturing methods affect the feature extraction performance, which reveals the cumulative effects on tracking and the trajectory path results.

**Dataset.** In the automotive visual odometry experiment, we collected images taken from a vehicle driving through a campus. To validate the performance under various illumination conditions, we tried to drive almost the same path three times, at 14:00, 16:00, and 18:00 hours. The exposure parameter for ME was initialized at 14:00. We chose a homing path to easily measure the translation error between the starting and ending points of the path. This error measurement has been used to evaluate many localization methods [49, 50] because only a few reference points are needed rather than the ground truth for the entire path.

**Visual odometry.** To conduct visual odometry, we calibrated the camera's intrinsic parameters using the method of [51] and calibrated the pitch angle and height from the ground using the vanishing points of an image. We used the monocular SLAM algorithm [45] for this experiment.

Figure 15 shows the visual odometry trajectories using representative source images from AE, ME, and Ours-single. Because Ours-single method captures richer gradient images, the subsequent algorithm extracts better features for tracking and performs more consistent localization with the smallest end-point errors across different times.

Images captured at two locations (P1 and P2) along the path are shown in Figure 16. We can observe that the illumination conditions varied widely across both space and time. Neither the AE nor the ME methods were able to successfully support feature tracking due to under-exposures. In contrast, our method captured well-exposed images and successfully supported the extraction of image features conducive to tracking.

In Figure 17, we present the quantitative evaluation results. The statistics of the number of inlier features are presented in (a), and the distance error ratios of each trajectory are presented in (b). Distance error ratio is computed by dividing the distance error between the starting point and ending point of a closed-loop trajectory by the length of an estimated trajectory. The distance error ratio is used as an evaluation metric because an estimated trajectory has a scale ambiguity in monocular visual odometry. In the figure, the results of our method consistently extracted a larger number of inlier features and resulted in smaller errors than did the AE and ME methods. For robust estimation, a larger number of inlier features is preferred, which is the main reason that our method achieved better results than the others.

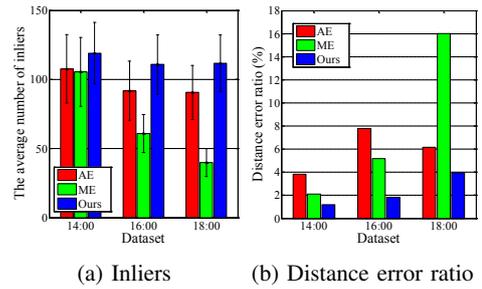

(a) Inliers  (b) Distance error ratio

Fig. 17: Quantitative evaluation results of the visual odometry experiment: (a) means and standard deviations of the number of inlier features, (b) relative distance errors.

### C. Multi-camera experiments

To evaluate Ours-multi and demonstrate the versatility of our proposed method, we performed experiments with three applications: surround-view imaging, panoramic imaging, and stereo matching.

*1) Implementation:* Figure 10-(b) shows our multi-camera experiment system with a mobile platform, the Clearpath Husky-A200. The multi-camera system has four cameras synchronized by a hardware trigger, which generates synchronization signals at a frequency of 20Hz. For each camera, we used a Sony ICX445 CCD $1/3''$ sensor with a fixed focal length of $1.4mm$. Thus, each camera had a $1288 \times 964$ resolution with a $58.44$ dB dynamic range and a $185° \times 144°$ field-of-view (FoV). The multi-camera system was fully calibrated, including both intrinsic and extrinsic parameters. To achieve the desired exposure level, we first adjusted the shutter speed until it reached the pre-defined maximum value. Then, we continued by the gain—the same approach as was used in the previous single-camera experiments.

In the subsequent experiments, the maximum shutter speed was set to $49ms$, and both the shutter speed and gain parameters of each camera were adjusted by the proposed nonlinear update function Eq. (4).

*2) Surround-view imaging application:* Surround-view imaging (Top-view) application is one of the most popular vision techniques in the automobile industry. A surround-view image for a vehicle is commonly created by stitching the ground parts of images from multiple cameras together; these images are mainly used to assist drivers while parking.

In this experiment, we generated a surround-view image from four partially overlapped input images as shown in

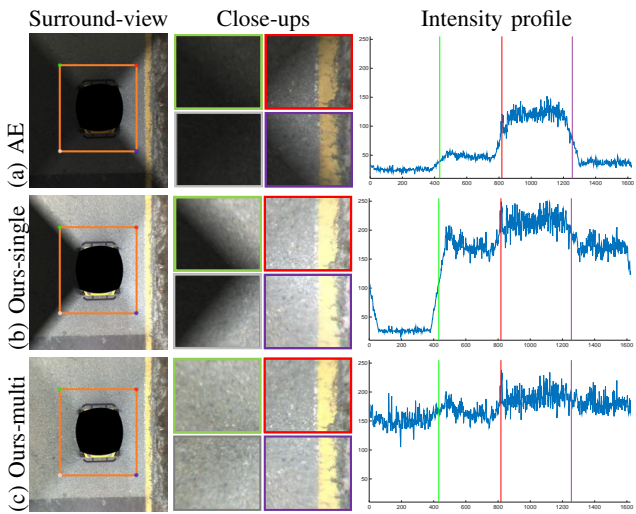

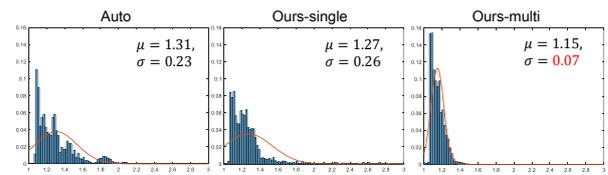

Fig. 18: A qualitative comparison of the surround-view application using different exposure methods: (a) AE, (b) Ours-single, and (c) Ours-multi. The image patches in the second column indicate regions in the vicinity of the green, red, purple, and gray points in the surround-view images. The last column shows the 1D signals of the pixel values on the orange lines of the surround-view images. We recommend that readers zoom-in to see the details clearly.

Fig. 19: Histogram of the intensity ratios for all overlapped regions

Figure 18. We pre-calibrated the relative poses between cameras using a flat-ground assumption and applied a simple linear blending method [7] as a post-processing step.. These overlapped regions are used to compute the relative brightness ratio $r^{ij}$ in Eq. (5) to balance the exposure across cameras. In this application, we consider only the bottom-half region of the original image domain (corresponding to the ground region) because only the ground regions of images are displayed to users. We compute $r^{ij}$ over the bottom half region.

Figure 18 shows a comparison of the surround-view image results from different exposure control algorithms. In the figure, we show the intensity changes at the boundary regions between cameras in the close-up views and also visualize the effect of exposure balancing by plotting the intensity profiles along the boundaries of the rectangle (orange colored lines) in the surround-view images. The colors of the boundaries in the close-up views correspond to the pixel locations with the same color points in the surround-view images, respectively. As shown in (a) and (b), the surround-view images processed by the AE and Ours-single methods have noticeably large intensity transitions at the boundary regions of two neighboring images; these are caused by the independent exposure control of each camera. In this condition, it is difficult to achieve constant brightness across all the cameras. In contrast, the Ours-multi method results in relatively small intensity transitions by virtue of our exposure balancing method.

In Figure 18-(c), the un-warped images of Ours-multi are over-saturated except in the ground regions. Note that we adjust camera exposure levels by considering only the ground regions because surround-view imaging is only concerned with the ground portions of the images (as shown in the surround-view images).

For the quantitative evaluation, we captured image sequences using a mobile robot with a multi-camera system under various illumination conditions (sunlight, cloudy, parking lot, and nighttime scene). We drove along almost the same path three times using different methods (AE, Ours-single, and Ours-multi) to obtain images for each dataset. Figure 19 presents the histograms of the intensity similarities of all the overlapped regions. These normalized histograms were computed by the intensity ratios between images from neighboring cameras in the overlapped region for all candidates. The histograms also plot an estimate of the probability density function for intensity ratio. The result obtained by Ours-multi had the smallest mean value and standard deviation of the intensity ratio, which means that the cameras controlled by Ours-multi result in the most similar images in the intensity space. This helps generate more natural surround-view images without post-processing steps, such as color transfer [52]. Additional results from the surround-view imaging experiment can be found in the supplementary material.

*3) Panoramic imaging application:* Panoramic imaging is another good example to demonstrate the effectiveness of our method for multi-camera exposure control. Panoramic imaging is similar to the previous surround-view imaging application; however, rather than only the ground portion, this application considers the entire region of images. Thus, it is important to be able to capture informative image features from scenes with an extremely wide dynamic range.

In this experiment, we used a cylindrical panoramic model [53] with simple alpha and linear blending methods [7]. Additionally, to accurately match each pair of images, we initialized the camera poses using pre-calibrated extrinsic parameters. Then, we optimized the initial camera poses by minimizing the re-projection error of the matched feature points in an overlapped region of each image pair. We used the feature matching framework in [54] with the Harris corner detector [55] and BRIEF feature descriptor [56]. The patches of corresponding feature points were used to compute the relative brightness ratios $r^{ij}$ in Eq. (5).

Figure 20 shows the results of panoramic image stitching. The panoramic images with simple alpha blending clearly show the differences between those resulting from exposure calculated by Ours-multi and those calculated by the conventional AE method. The strong sunlight behind the building causes a wide dynamic range in the scene; therefore, the images obtained by the AE method have large brightness differences, and many informative parts of the foreground region are under-saturated. Such wide difference can cause the following computer vision algorithms, which mainly operate on foreground objects, to fail. In contrast, the results produced by Ours-multi show small intensity gaps in the overlapped regions and preserve the important information in the foreground regions well. By smoothing the brightness differences between the overlapped regions, the linear blending operation causes both results to be visually pleasing, but it is unable to recover information already lost in the foreground region as shown in the close-up views in Figure 20 (a).

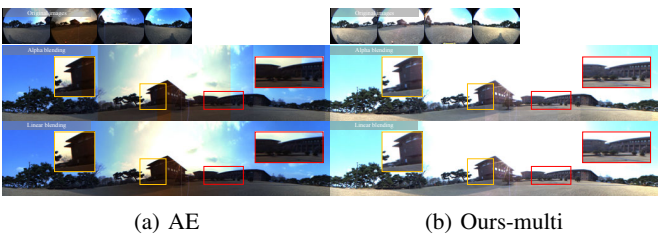

(a) AE  (b) Ours-multi

Fig. 20: Examples of panoramic images with alpha and linear blending. The top images of both (a) and (b) were generated by alpha blending ($\alpha = 0.5$), and the bottom images were generated by linear blending.

*4) Stereo matching application:* In the previous two experiments, we compared differences among exposure control methods by showing the results in the color image domain. In this experiment, we applied stereo matching algorithms to verify the effectiveness of the Ours-multi method in the gradient domain. Following the panoramic imaging experiment, we used the feature matching framework in [54] and computed the relative brightness ratios $r^{ij}$ in Eq. (5) using corresponding patches. For this experiment, we applied two representative stereo matching methods: area-correction stereo (Block, [57]) and semi-global matching (SGM, [58]).

Figure 21 shows the results of stereo matching. The block-matching method computes disparity by comparing the sum of absolute differences (SAD) in a local area without any pre- or post-processing and regularization. This approach allows us to directly compare the quality of images that affect the patch-matching performance. In the images obtained by the AE method, much of the visual information in the foreground regions is missing because of the strong illumination behind the foreground. In particular, it is difficult to recognize where and what the pedestrian in (b) and the bush in (c) are in the block matching results. In contrast, Ours-multi captures the important image features in the foreground regions; consequently, its disparity map quality is better than that of the AE.

SGM improves the quality of the disparity maps for both the AE and Ours-multi methods by virtue of its spatial regularization. However, a visible gap still exists between the AE and Ours-multi methods even after optimization. Ours-multi presents clearer and more dense disparity maps than does the AE method. This result further highlights the importance of image quality at the image-capturing stage; difficulties during image capture may cause subsequent algorithms to fail.

## VI. DISCUSSION AND CONCLUSIONS

In this paper, we presented a novel auto-exposure method that is designed specifically for gradient-based vision systems

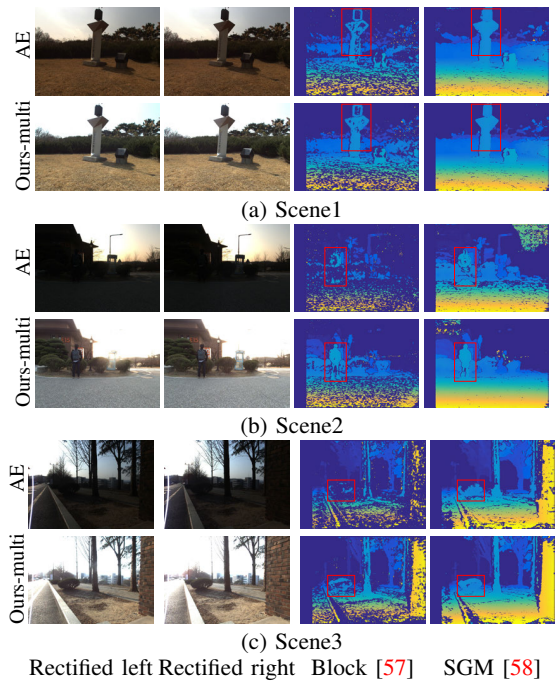

(a) Scene1

(b) Scene2

(c) Scene3

Rectified left  Rectified right  Block [57]  SGM [58]

Fig. 21: Qualitative comparison of estimated disparity maps from two exposure methods: AE and Ours-multi. The images of the first two columns are rectified stereo image pairs, and the third and fourth columns show the estimated disparity maps using Block [57] and SGM [58], respectively.

on outdoor mobile platforms. In outdoor environments, scene radiance often has a much wider dynamic range than the possible range of cameras. The conventional camera exposure calculation methods fail to capture well-exposed images, which degrades the performance degradation of subsequent image-processing algorithms. To solve this problem, our auto-exposure method adjusts the camera exposure to maximize the gradient information with respect to the current scene; therefore, our method is able to determine a more appropriate exposure that is robust to severe illumination changes. Moreover, our proposed exposure balancing method is extended to multi-camera systems.

We evaluated our methods through extensive outdoor experiments with off-the-shelf machine vision cameras using a variety of image processing and computer vision applications. We believe that our method constitutes an alternative or additive solution to conventional auto-exposure methods, especially for outdoor computer vision systems used for purposes such as surveillance monitoring and autonomous mobile platforms.

We close this paper with a further discussion of our method and its limitations as follows.

**Discussion and Limitation** Compared to the conventional AE methods, we have shown that the proposed method is especially beneficial when capturing scenes that involve dark foreground regions and strong back-lighting are involved (*i.e.*, scenes with wide dynamic range). Note that this does not suggest that our method is a complete alternative solution to the conventional AE and ME methods, which have been used in enormous numbers of situations. When a scene has mild light conditions, conventional auto-exposure methods are already a good choice. In this regard, the value of the proposed method is that it provides a strong alternative approach that can be beneficially selected and used depending on the user scenario, such as in outdoor computer vision systems.

In several of the figures, some of the images obtained by our method appear brighter than the images obtained by AE. This occurs because AE adjusts camera exposure by directly measuring brightness values, while our method adjusts it only by maximizing the gradient information. Hence, this difference in brightness is not an issue from the perspective of computer vision applications as long as the exposure captures the dynamic range of the foreground well and the brightness consistency between multi-view images is preserved.

When a scene has different objects with different textures, our method may overlook the parts of objects whose texture gradient is relatively lower than that of others. While this results from a reasonable decision by the algorithm to maximize the holistic gradient information in an image, it can lead to some failure cases. This limitation might be addressed by considering additional semantic information in scenes, which will be interesting future work.

In the panoramic imaging and stereo matching applications of multi-camera systems, Ours-multi requires that the FoVs of cameras overlap to create matching corresponding patches. In addition, our method depends on the performance of the patch matching algorithms, which typically exploit the consistency of patch brightness. Therefore, we can adjust the balance parameter $\alpha$ to encourage brightness consistency between images to improve the matching performance. In this paper, we showed that this approach leads to more stable matching than does independent per-camera AE. Another approach is to use an advanced intensity-invariant feature; however, doing so is out of the scope of this work. Note that the problem of patch-matching is not an issue when the matching patches can be obtained through pre-calibration, *e.g.*, the surround-view imaging case.

Along the same lines, the gradient metric approach used here may not always be optimal because it is hand-designed. A learning-based metric is another promising future direction.

Computational cost is a crucial issue in camera exposure control. Most computation is conducted through pixel-by-pixel operations of independent processes such as gradient information metric computation, gamma-mapped image generation, and downsampling—all of which parallelizable. However, a memory bottleneck occurs when generating various gamma-mapped images. As shown in Fig. 4-(b,c), because we can use lower-resolution images to facilitate faster computation that uses less memory without incurring a significant loss in accuracy, the memory bottleneck can be adjusted accordingly. This feature of our method would be useful if it were implemented as an embedded system.

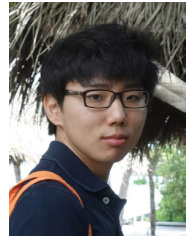

**Inwook Shim** received the B.S degree in Computer Science from Hanyang University in 2009, the M.S degree in Robotics Program from KAIST in 2011, and Ph.D degrees in Division of Future Vehicle from KAIST in 2017. He is currently a research scientist at the Agency for Defense Development. He was a member of "Team KAIST" that won the first place at DARPA Robotics Challenge Finals in 2015. He was a recipient of Qualcomm Innovation Award, NI finalist of NI Student Design Showcase, and received the KAIST achievement award of Robotics, and Creativity and Challenge award from KAIST. His research interests include robot vision for autonomous systems and deep learning.

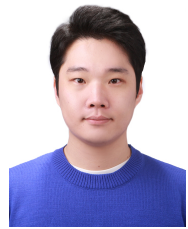

**Tae-Hyun Oh** is a postdoctoral associate at MIT/CSAIL. He received the BE degree (The highest ranking) in Computer Engineering from Kwang-Woon University, South Korea in 2010, and the MS and PhD degrees in Electrical Engineering from KAIST, South Korea in 2012 and 2017, respectively. He was a visiting student in the Visual Computing group, Microsoft Research, Beijing and in the Cognitive group, Microsoft Research, Redmond in 2014 and 2016, respectively. He was a recipient of Microsoft Research Asia fellowship, a gold prize of Samsung HumanTech thesis award, Qualcomm Innovation award and a top research achievement award from KAIST.

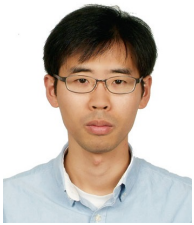

**Joon-Young Lee** received the B.S degree in Electrical and Electronic Engineering from Yonsei University, Korea in 2008. He received the M.S and Ph.D degrees in Electrical Engineering from KAIST, Korea in 2009 and 2015, respectively. He is currently a research scientist at Adobe Research. His research interests include deep learning, photometric image analysis, image enhancement, and computational photography. He is a member of the IEEE.

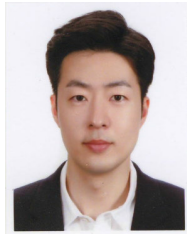

**Jinwook Choi** received the B.S. and Ph.D. degrees from the School of Electrical and Electronic Engineering, Yonsei University, Seoul, Korea, in 2008 and 2014, respectively. He is currently a Senior Research Engineer with the ADAS Development Team of Automotive Research and Development Division, Hyundai Motor Company, Korea. His research interests include 2D/3D image and video processing, computer vision, hybrid sensor systems, and intelligent vehicle systems.

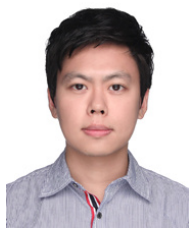

**Dong-Geol Choi** received the B.S and M.S degree in Electric Engineering and Computer Science from Hanyang University in 2005 and 2007, respectively, and the Ph.D degrees in the Robotics Program from KAIST in 2016. He is currently a post-doctoral researcher at the Information & Electronics Research Institute, in KAIST. His research interests include sensor fusion, autonomous robotics, and artificial intelligence issues. Dr.Choi received a fellowship award from Qualcomme Korea R&D Center in 2013. He was a member of "Team KIAST," which won the first place in DARPA Robotics Challenge Finals 2015.

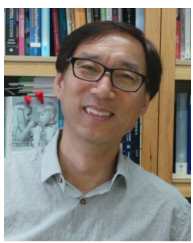

**In So Kweon** received the B.S. and M.S. degrees in Mechanical Design and Production Engineering from Seoul National University, South Korea, in 1981 and 1983, respectively, and the Ph.D. degree in Robotics from the Robotics Institute, Carnegie Mellon University, USA, in 1990. He worked for the Toshiba R&D Center, Japan, and he is now a KEPCO chair professor with the Department of Electrical Engineering, since 1992. He served as a program co-chair for ACCV 07' and a general chair for ACCV 12'. He is on the honorary board of IJCV. He was a member of 'Team KAIST' which won the first place in DARPA Robotics Challenge Finals 2015. He is a member of the IEEE and the KROS.